\documentclass[10pt,twocolumn,letterpaper]{article}
\usepackage[pagenumbers]{cvpr}

\usepackage{amsmath,amssymb,booktabs,graphicx,microtype}

\usepackage{float}
\definecolor{cvprblue}{rgb}{0.21,0.49,0.74}
\usepackage[breaklinks,colorlinks,allcolors=cvprblue]{hyperref}
\hypersetup{pdftitle={RAMP: Rethinking Multimodal Evaluation through Answer Readout Protocols},pdfauthor={Jianye Zeng, Yikang Qiao, Duan Huang, Ziying Song}}
\newcommand{\pp}{\,\mathrm{pp}}
\newcommand{\ramp}{{\upshape\textsc{RAMP}}}
\newcommand{\bare}{\textit{bare}}
\newcommand{\surf}{\textit{surface-max}}
\newcommand{\typed}{\textit{typed}}
\newcommand{\rev}{\textit{typed-reversed}}
\newcommand{\Yes}{\mathrm{yes}}
\newcommand{\No}{\mathrm{no}}
\title{RAMP: Rethinking Multimodal Evaluation through Answer Readout Protocols}
\author{
Jianye Zeng$^{1,*}$ \quad
Yikang Qiao$^{2,*}$ \quad
Duan Huang$^{2,\dagger}$ \quad
Ziying Song$^{3,\dagger}$ \\[0.45em]
{\footnotesize $^1$University of Science and Technology of China \quad $^2$Central South University \quad $^3$Yanshan University} \\[0.25em]
{\footnotesize $^*$These authors contributed equally to this work. \qquad
$^\dagger$Corresponding authors.}
}
\begin{document}
\setlength{\textfloatsep}{10pt plus 2pt minus 2pt}
\setlength{\dbltextfloatsep}{10pt plus 2pt minus 2pt}
\setcounter{dbltopnumber}{2}
\renewcommand{\dbltopfraction}{0.92}
\renewcommand{\textfraction}{0.08}
\maketitle
\vspace{-12pt}

\begin{abstract}
Multimodal large language models (MLLMs) are often evaluated by converting model outputs into discrete answers. The conversion protocol determines which scores become predictions and can strongly influence the reported results. We introduce \ramp{} (Readout Aware Multimodal Protocol), a systematic framework that compares direct yes/no token logits, surface-form scores, and typed or reversed A/B interfaces under fixed model checkpoints and paired visual questions. On 500 POPE adversarial questions, typed versus bare readout changes accuracy by $-20.2$ percentage points for SmolVLM-256M, $-16.4$ for TinyLLaVA-1.1B, and $+1.8$ for Qwen2.5-VL-3B. Paired transitions reveal extensive prediction changes even when the net accuracy difference is small: 79 of 104 binary ChartQA items change correctness for SmolVLM-500M, despite a $+1.0$-point net difference. We further analyze prompt wording, candidate forms, option position, cross-task effects, and calibration. An independent offline replay reproduces 16,140 first-token decisions from saved raw scores with exact agreement. Our findings establish answer readout as a measurable component of multimodal evaluation and provide a reproducible protocol for analyzing model comparisons.
\end{abstract}


\begin{figure}[!t]
  \centering
  \includegraphics[width=\columnwidth]{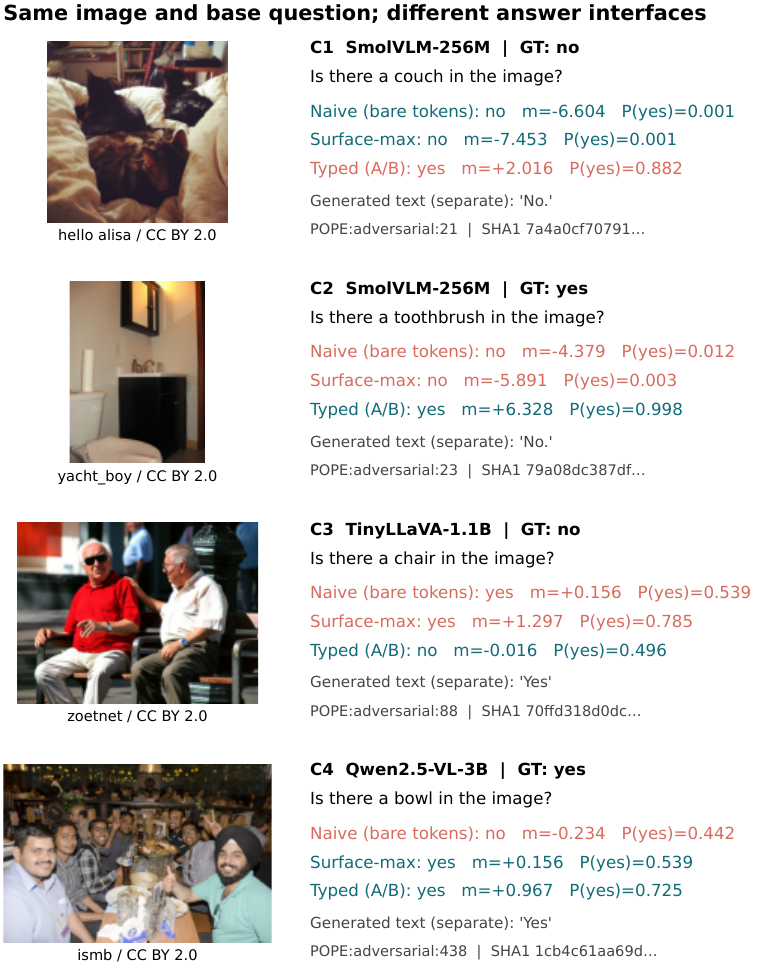}
  \caption{\textbf{Readout-sensitive examples.} Each row fixes the image and base question while changing only the answer interface. Bare-token, surface-form, and typed A/B readouts can produce different predictions and score margins on the same item; here $m$ denotes the signed yes--no score margin and $P(\mathrm{yes})=\sigma(m)$.}
  \label{fig:teaser}
\end{figure}

\begin{figure*}[!t]
\centering\includegraphics[width=\textwidth]{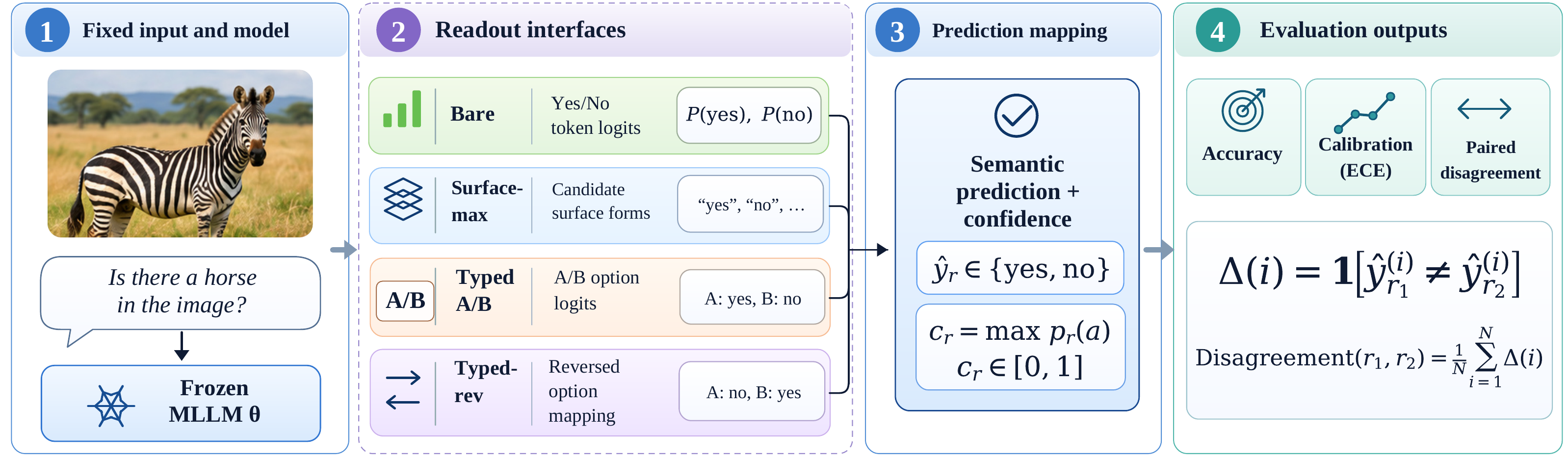}
\caption{\textbf{Overview of RAMP.} A fixed image--question pair is evaluated with frozen MLLM checkpoint. Bare and surface form scoring inspect answer-token logits under the original question. Typed and reversed interfaces score option letters after independently constructed A/B prompts. Semantic mapping converts each candidate decision to a shared yes/no label. Paired analysis compares accuracy, transitions, and confidence.}
\label{fig:protocol}
\end{figure*}

\section{Introduction}\label{sec:intro}
Multimodal large language models (MLLMs) have become a common foundation for image-grounded dialogue and visual question answering. Visual instruction tuning equips language models with flexible image-conditioned response interfaces \cite{liu2023llava}. The original visual question answering benchmark introduced open-ended and multiple-choice answer formats \cite{antol2015vqa}. Subsequent benchmarks have expanded evaluation toward compositional understanding, object hallucination, chart reasoning, and specialized multimodal knowledge \cite{li2023pope}. 

An evaluation score emerges from several interacting components. The image and question define the task; the model produces scores or a sequence of tokens; the evaluation program converts those outputs into a benchmark answer. This last operation is the \emph{answer readout}. It specifies which answer tokens are considered, how multiple candidate forms are combined, and how generated responses are mapped to labels. The readout is particularly consequential for binary visual questions, where different interfaces can express the same semantic choice with very different token sequences.

Answer interfaces influence both discrete predictions and their reported confidence. Standard calibration compares predicted confidence with an explicitly defined event \cite{guo2017calibration}. For a binary visual question, the probability assigned to \texttt{yes} and the confidence that the selected answer is correct describe separate events. Language model question answering has demonstrated the importance of carefully defining confidence estimates \cite{jiang2021know}. Prompt wording can also shift language model decisions even when the evaluation examples remain unchanged \cite{zhao2021calibrate}. 

However, a useful measurement must account for differences at the individual-question level. Two readout configurations can report nearly identical accuracy while succeeding on different examples. Correct-to-wrong and wrong-to-correct transitions cancel in aggregate accuracy; the number of changing decisions records a complementary form of sensitivity.  These factors motivate a paired evaluation in which the model checkpoint, data membership, and extraction rules are explicitly recorded for every condition.

We introduce \ramp{}, a readout-aware framework for controlled multimodal evaluation. It represents an answer interface through its prompt, candidate token inventory, semantic mapping, and deterministic decision rule. It combines aggregate accuracy with paired correctness transitions, confidence calibration, and image-cluster uncertainty intervals. The framework supports transparent comparisons among direct logits, surface forms, and typed answer interfaces on the same visual questions.

Our experiments span three primary MLLM backbones on 500 POPE adversarial questions, an additional SmolVLM-500M control, and a 104-question native binary subset of ChartQA. We also investigate prompt and token forms, candidate-set expansions, option position, and device configurations. 

\noindent\textbf{Our contributions are:}
\begin{itemize}\setlength{\itemsep}{1pt}
\item We formulate answer readout as an explicit multimodal evaluation protocol, covering token-level candidate scoring, semantic label mappings, typed options, and paired counterfactual interfaces.
\item We develop a reproducible analysis of readout sensitivity using accuracy differences, item level correctness transitions, calibration diagnostics, and image-cluster bootstrap uncertainty.
\item We provide systematic evidence across POPE, binary ChartQA, and targeted interface controls, supported by an independent replay of 16,140 saved first-token predictions.
\end{itemize}

\section{Related Work}\label{sec:related}
\textbf{Multimodal evaluation.} VQA v2 reduces question-only answer priors and emphasizes visual evidence \cite{goyal2017vqa}. POPE evaluates object hallucination through structured binary polling \cite{li2023pope}. MME provides perception and cognition tests \cite{fu2023mme}. HallusionBench targets visual illusion and language hallucination \cite{guan2024hallusion}. ChartQA examines chart-based visual and numerical reasoning \cite{masry2022chartqa}. TextVQA focuses on reading scene text \cite{singh2019textvqa}. MMMU covers expert-level multimodal subjects \cite{yue2024mmmu}. MM-Vet studies integrated capabilities \cite{yu2024mmvet}. MMBench evaluates broad vision-language reasoning \cite{liu2024mmbench}. SEED-Bench provides systematic multimodal multiple-choice evaluation \cite{li2024seed}. These benchmarks employ diverse response formats and motivate a common treatment of answer extraction.

\textbf{Multimodal model interfaces.} InstructBLIP adapts visual features to natural-language instructions \cite{dai2023instructblip}. BLIP-2 connects visual encoders to frozen language models \cite{li2023blip2}. Few-shot language-model behavior can depend on example order \cite{lu2022order}. We study TinyLLaVA \cite{zhou2024tinyllava}, Qwen2.5-VL \cite{bai2025qwen}, and SmolVLM checkpoints \cite{marafioti2025smol} using controlled answer interfaces rather than architectural modifications.

\textbf{Confidence and statistical evaluation.} Calibration connects reported confidence to empirical correctness \cite{guo2017calibration}. Proper scoring rules provide a probabilistic basis for evaluation \cite{gneiting2007scoring}. Paired bootstrap methods quantify uncertainty over sampled items and clusters \cite{efron1993bootstrap}. \ramp{} combines these ideas with readout-specific margins, explicit target events, and matched question-level comparisons.

\section{Method: Readout-Aware Evaluation}\label{sec:method}
\subsection{Problem formulation}
Let $\mathcal D=\{(x_i,q_i,y_i,h_i)\}_{i=1}^{N}$ contain an image $x_i$, a question $q_i$, a gold semantic label $y_i\in\mathcal Y=\{\Yes,\No\}$, and an image identity $h_i$. The image identity groups questions derived from the same source image. A pretrained MLLM $f_\theta$ produces logits at the first answer-token position. The parameters $\theta$, visual samples, and semantic labels are held fixed within a matched comparison.

We represent the answer interface by a protocol tuple
\begin{equation}\label{eq:tuple}
 r=(t_r,C_r,g_r,\tau_r),\qquad r\in\mathcal R,
\end{equation}
where $t_r$ constructs the prompt and answer prefix, $C_r$ specifies admissible single-token candidates, $g_r:C_r\rightarrow\mathcal Y$ maps candidates to semantic answers, and $\tau_r$ defines deterministic tie handling. The context for item $i$ is $u_{ir}=t_r(x_i,q_i)$. For candidate token $v$, the extracted score is
\begin{equation}\label{eq:logits}
 \ell_{ir}(v)=\operatorname{logit}_{\theta}(v\mid u_{ir}),\quad v\in C_r.
\end{equation}
Candidate scores at a common output position share the same softmax normalization, so logit differences directly determine the corresponding restricted candidate probabilities.

\subsection{Semantic candidate scores and decisions}
Several token spellings may express the same label. \ramp{} associates each semantic answer with its highest-scoring available single-token form:
\begin{align}\label{eq:decision}
 S_{ir}(y)&=\max_{v\in C_r:g_r(v)=y}\ell_{ir}(v),\\
 \hat y_{ir}&=\mathop{\operatorname{argmax}}_{y\in\mathcal Y} S_{ir}(y).\nonumber
\end{align}
The implementation breaks equal-score ties toward \texttt{yes}. Candidate inventory and token IDs are retained for every run. The maximum in Eq.~\ref{eq:decision} defines a reproducible, first-token score aggregation rule. It supports a direct comparison of readout conventions across recorded tokenizer vocabularies.

The signed semantic margin summarizes the decision boundary:
\begin{equation}\label{eq:margin}
 m_{ir}=S_{ir}(\Yes)-S_{ir}(\No).
\end{equation}
A positive margin selects \texttt{yes}; a negative margin selects \texttt{no}. Its magnitude measures the separation between semantic candidates under the selected interface. The margin is especially useful for analyzing items whose decisions change when a prompt or candidate inventory is modified.

\subsection{Answer-readout families}
\textbf{Bare-token readout.} The \bare{} condition compares the original lowercase \texttt{yes} and \texttt{no} candidate logits under the benchmark prompt. It is equivalent to the \textit{naive} first-token condition in the archived results. Its prompt template is denoted $t_0$ and its candidate set is $C_{\mathrm{bare}}=\{v_{\Yes},v_{\No}\}$.

\textbf{Surface-form readout.} The \surf{} condition uses $t_0$ and expands the candidate inventory to the available single-token surface forms of both labels. Writing $V_y$ for the available forms of semantic label $y$, its score is
\begin{equation}\label{eq:surface}
 S_{i,\mathrm{surface}}(y)=\max_{v\in V_y\cap C_{\mathrm{surface}}}\ell_{i0}(v).
\end{equation}
This construction measures the influence of capitalization and token boundary conventions while keeping the original prompt fixed.

\textbf{Typed option readout.} The \typed{} interface presents \texttt{A. yes} and \texttt{B. no} before the answer prefix. With letter scores $\ell_{i,+}$, the mapping is $g_{+}(\mathrm A)=\Yes$ and $g_{+}(\mathrm B)=\No$. The \rev{} interface constructs a new prompt with the options reversed:
\begin{align}\label{eq:ab}
 m_{i,+}&=\ell_{i,+}(\mathrm A)-\ell_{i,+}(\mathrm B),\\
 m_{i,-}&=\ell_{i,-}(\mathrm B)-\ell_{i,-}(\mathrm A).\nonumber
\end{align}
Both margins use the same semantic orientation. Each option mapping is scored under its own prompt context, which captures the joint effect of answer wording, option letters, and displayed order.

A generation-based interface can be defined with a decoder $G_\theta(u_i)$ and a response parser $\pi$ through $\hat y_i^{\mathrm{gen}}=\pi(G_\theta(u_i))$. The present experiments concentrate on the four logged first-token interfaces above. Stored generated strings are retained as a separate analysis input when available.

\subsection{Paired accuracy and disagreement}
For item $i$, define correctness $z_{ir}=\mathbf 1[\hat y_{ir}=y_i]$. Given protocols $a$ and $b$ on the same $N$ items,
\begin{align}\label{eq:acc}
 \mathrm{Acc}(r)&=\frac{1}{N}\sum_{i=1}^{N}z_{ir},\\
 \Delta_{b,a}&=\mathrm{Acc}(b)-\mathrm{Acc}(a).\nonumber
\end{align}
Let $n_{01}=\#\{i:z_{ia}=0,z_{ib}=1\}$ and $n_{10}=\#\{i:z_{ia}=1,z_{ib}=0\}$. The paired effect decomposes as
\begin{equation}\label{eq:decomp}
 \Delta_{b,a}=\frac{n_{01}-n_{10}}{N},\quad
 d_{b,a}=\frac{n_{01}+n_{10}}{N}.
\end{equation}
Here $d_{b,a}$ is the correctness-transition rate. A complementary semantic flip rate is $f_{b,a}=N^{-1}\sum_i\mathbf 1[\hat y_{ia}\ne\hat y_{ib}]$. On the strictly binary questions in this paper, semantic changes correspond to correctness transitions. Equation~\ref{eq:decomp} reveals the cancellation hidden by a small aggregate difference.

\subsection{Readout confidence and calibration}
We normalize the two semantic scores to obtain the readout-specific probability proxy
\begin{equation}\label{eq:conf}
 p_{ir}=\frac{e^{S_{ir}(\Yes)}}{e^{S_{ir}(\Yes)}+e^{S_{ir}(\No)}}
 =\sigma(m_{ir}),\quad c_{ir}=\max(p_{ir},1-p_{ir}).
\end{equation}
The confidence target for correctness is $z_{ir}$. With evaluation bins $\mathcal B_1,\ldots,\mathcal B_B$, the corresponding expected calibration error is
\begin{equation}\label{eq:ece}
 \mathrm{ECE}_r=\sum_{b=1}^{B}\frac{|\mathcal B_b|}{N}
 \left|\overline c_b-\overline z_b\right|.
\end{equation}
We use the 15-bin scheme documented for each experiment export and evaluate correctness AUROC by ranking $z_{ir}$ with $|m_{ir}|$. Positive-label calibration instead uses $p_{ir}$ and target $\mathbf 1[y_i=\Yes]$. Naming the target preserves the interpretation of each statistic.

\subsection{Paired uncertainty and auditability}
We resample distinct image identities $h_i$ with replacement. All questions belonging to a sampled image are retained together. For $B_{\mathrm{boot}}=1{,}000$ image-cluster bootstrap draws, the paired interval is
\begin{equation}\label{eq:boot}
 \mathrm{CI}_{95\%}(\Delta)=
 [Q_{0.025}(\Delta^{(b)}),Q_{0.975}(\Delta^{(b)})].
\end{equation}
The audit record includes model and processor revisions, rendered prompts, candidate token IDs, device and precision, sample/image identities, raw logits, and semantic mappings. These fields permit score to answer reconstruction and sample level paired analyses.

\section{Experimental Setup}\label{sec:setup}
\subsection{Models and benchmark configurations}
We evaluate the recorded checkpoints of SmolVLM-256M, TinyLLaVA-1.1B, and Qwen2.5-VL-3B on 500 adversarial POPE questions. The primary new runs use MPS/fp16 and checkpoint-specific documented preprocessing. SmolVLM-500M is evaluated as an additional CPU/fp32 single-tile configuration. The cross-task study uses 104 native yes/no questions from ChartQA human-test, containing 54 positive and 50 negative labels over 99 distinct image hashes. Every condition compares matched samples within its recorded run identity.

\subsection{Evaluation matrix and saved evidence}
Four main readouts are evaluated: \bare{}, \surf{}, \typed{}, and \rev{}. Additional controlled runs vary prompt wording, available answer-token forms, nested candidate-set sizes $k\in\{2,5,20,50\}$, the position of A/B options, and the compute environment. We retain archived POPE results as a separately labeled series. Historical Qwen reversal uses its 487 matched observations; historical TinyLLaVA reversal has no saved paired arm.

The supplied processed bundle contains 57,073 readout-level predictions across 23 run identities. An accuracy replay recovers 217 exported accuracy cells. The raw JSONL candidate scores support an additional token-to-label replay of 16,140 decisions, with exact agreement against the corresponding processed predictions. Our results follow the saved run identities and distinguish historical from checkpoint-pinned evidence throughout.

\begin{table*}[!t]
\centering\small
\caption{\textbf{Readout accuracy under recorded model configurations.} Values are accuracy percentages on matched questions. Brackets give image-cluster 95\% confidence intervals for typed-minus-bare differences in percentage points. POPE SmolVLM-500M uses CPU/fp32 single-tile processing; the first three POPE rows use their separately recorded MPS/fp16 configurations. ChartQA uses 104 native yes/no questions.}
\label{tab:results}
\vspace{2pt}
\setlength{\tabcolsep}{6pt}
\begin{tabular}{llrrrrrl}
\toprule
Dataset / configuration & Model & $n$ & Bare & Surface & Typed & Reverse & Typed $-$ bare (pp) \\
\midrule
POPE, new MPS & SmolVLM-256M & 500 & 71.8 & 75.0 & 51.6 & 44.6 & $-20.2\;[-27.6,-13.2]$ \\
POPE, new MPS & TinyLLaVA-1.1B & 500 & 82.4 & 78.4 & 66.0 & 49.8 & $-16.4\;[-20.8,-11.8]$ \\
POPE, new MPS & Qwen2.5-VL-3B & 500 & 85.2 & 86.6 & 87.0 & 86.0 & $+1.8\;[-0.4,+4.0]$ \\
POPE, CPU/single tile & SmolVLM-500M & 500 & 82.2 & 81.2 & 60.0 & 60.4 & $-22.2\;[-27.4,-17.4]$ \\
\midrule
ChartQA binary & SmolVLM-256M & 104 & 48.1 & 50.0 & 54.8 & 48.1 & $+6.7\;[-6.9,+19.2]$ \\
ChartQA binary & SmolVLM-500M & 104 & 52.9 & 58.7 & 54.8 & 49.0 & $+1.9\;[-15.2,+17.5]$ \\
ChartQA binary & TinyLLaVA-1.1B & 104 & 56.7 & 53.8 & 52.9 & 52.9 & $-3.8\;[-13.6,+5.9]$ \\
ChartQA binary & Qwen2.5-VL-3B & 104 & 70.2 & 76.9 & 72.1 & 70.2 & $+1.9\;[-4.8,+8.7]$ \\
\bottomrule
\end{tabular}

\end{table*}

\begin{figure*}[!t]
\centering\includegraphics[width=.89\textwidth]{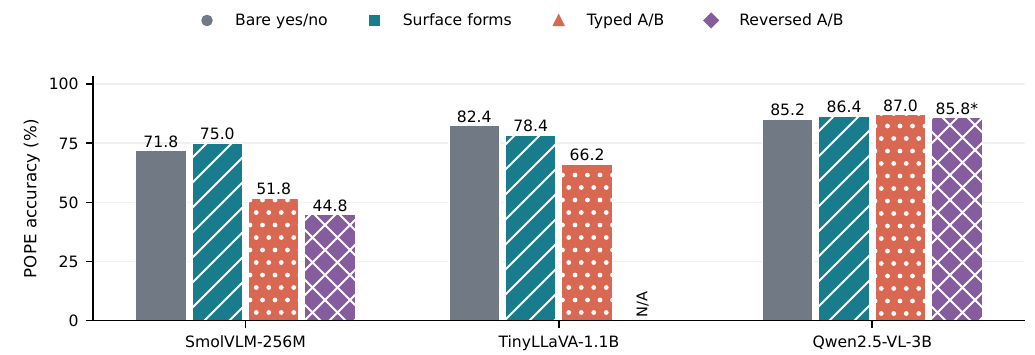}
\caption{\textbf{Archived POPE accuracy across answer readouts.} The saved historical runs evaluate bare yes/no, surface-form, typed A/B, and reversed A/B extraction on POPE adversarial questions. All displayed configurations use 500 items except Qwen's archived reversed arm (487 observed items); TinyLLaVA's reversed arm was not recorded. Archived results are separate from the checkpoint-pinned runs in Table~\ref{tab:results}.}
\label{fig:archived_accuracy}
\end{figure*}

\section{Results and Analysis}\label{sec:results}

\subsection{Answer readout changes measured accuracy}
Table~\ref{tab:results} reports all four readout interfaces in the pinned-checkpoint experiments. SmolVLM-256M drops from 71.8\% under bare scoring to 51.6\% under typed A/B scoring, an effect of $-20.2\pp$ with a 95\% paired interval of $[-27.6,-13.2]\pp$. TinyLLaVA-1.1B changes from 82.4\% to 66.0\%, corresponding to $-16.4\pp$ and interval $[-20.8,-11.8]\pp$. Qwen2.5-VL-3B moves from 85.2\% to 87.0\%, an estimated $+1.8\pp$ with interval $[-0.4,+4.0]\pp$. 

The surface-form condition provides a second perspective. SmolVLM-256M increases from 71.8\% to 75.0\% when the candidate set incorporates available spelling variants, whereas TinyLLaVA changes from 82.4\% to 78.4\%. Qwen rises from 85.2\% to 86.6\%. These comparisons share the original prompt and modify only the candidate scoring inventory. 

A separately configured SmolVLM-500M POPE run gives 82.2\% under bare readout and 60.0\% under typed readout, yielding $-22.2\pp$. This family control uses CPU/fp32 single-tile preprocessing and is reported by its complete configuration. The archived 500-item runs show the corresponding typed-minus-bare effects of $-20.0\pp$ for SmolVLM-256M, $-16.2\pp$ for TinyLLaVA, and $+1.8\pp$ for Qwen. Their recomputed image-cluster intervals are $[-27.2,-12.8]$, $[-20.6,-11.6]$, and $[-0.4,+4.0]\pp$, respectively.  Figure~\ref{fig:archived_accuracy} presents the complete archived four-interface comparison. The grouped view makes the contrast between surface-form and option-letter scoring directly visible: for SmolVLM-256M the archived surface condition reaches 75.0\% accuracy, while typed and reversed typed reach 51.8\% and 44.8\%. TinyLLaVA follows a similar direction, and Qwen retains high accuracy across its observed interfaces. The reversed Qwen bar is interpreted on its recorded 487-item subset.

\begin{figure}[!b]
\centering\includegraphics[width=\columnwidth]{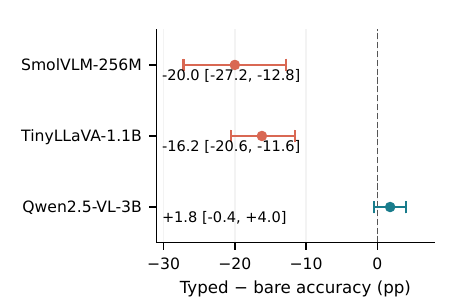}
\caption{\textbf{Paired accuracy changes in archived POPE runs.} Typed-minus-bare differences with image-cluster 95\% bootstrap intervals on 500 matched questions.}
\label{fig:paired}
\end{figure}

\begin{figure*}[!t]
\centering\includegraphics[width=.95\textwidth]{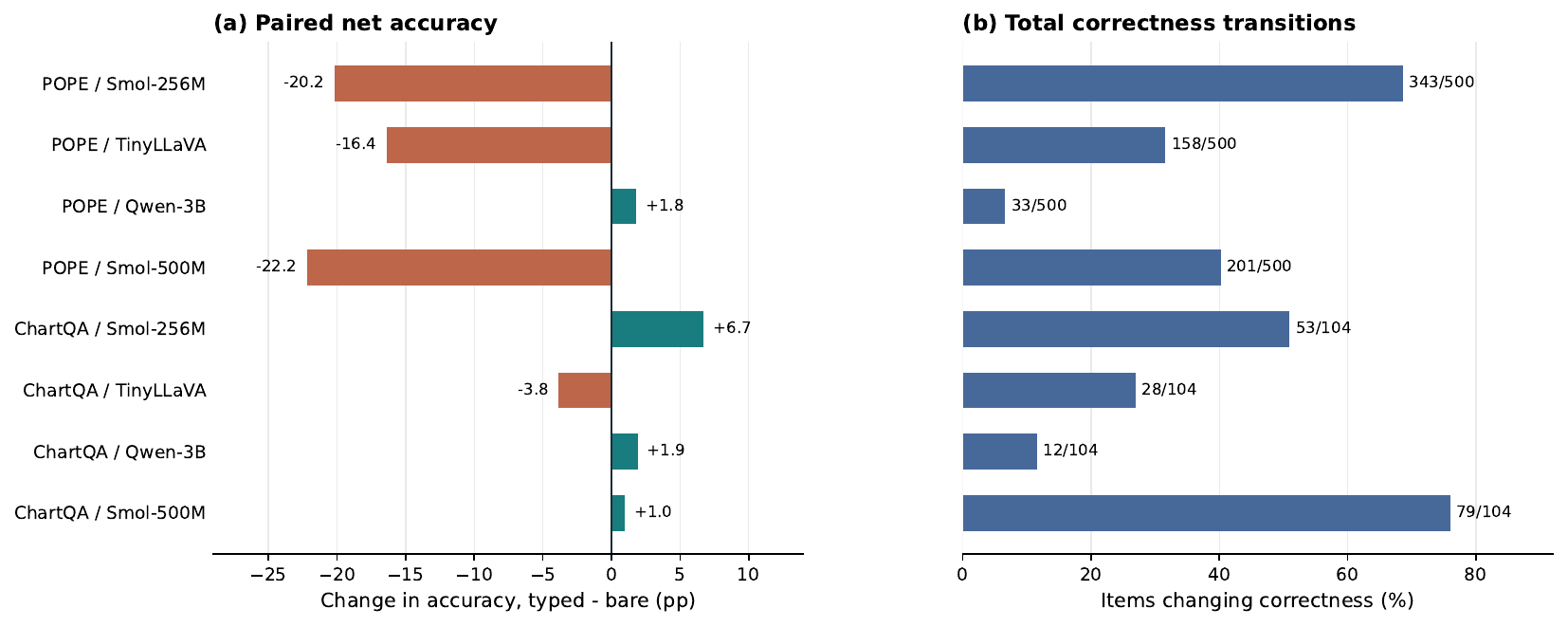}
\caption{\textbf{Paired readout effects reconstructed from raw scores.} Net typed-minus-bare accuracy changes (left) and the total number of correctness transitions (right) on matched items. The figure identifies the respective POPE and ChartQA run configurations. Large individual-question disagreement can accompany a small aggregate score difference.}
\label{fig:flips}
\end{figure*}

\subsection{Paired transitions reveal hidden disagreement}\label{sec:flips}
The decomposition in Eq.~\ref{eq:decomp} distinguishes net accuracy change from changes in the identity of correct predictions. In the archived SmolVLM-256M POPE run, switching from bare to typed moves 221 items from correct to incorrect and 121 from incorrect to correct. Thus 342 of 500 items change correctness, while the net effect is $-100/500=-20.0\pp$. TinyLLaVA records 119 correct-to-wrong and 38 wrong-to-correct transitions, giving 157 changed items and a $-16.2\pp$ net effect. Qwen records 12 and 21 transitions, respectively, yielding a $+1.8\pp$ net effect across 33 changed items. 

An especially informative case appears in the new ChartQA binary subset. SmolVLM-500M has 40 wrong-to-correct transitions and 39 correct-to-wrong transitions between bare and typed scoring. The 79 changed items represent 76.0\% of the 104-question subset. Their near cancellation produces an accuracy difference of only $+1/104$, approximately $+1.0\pp$. Figure~\ref{fig:flips} visualizes net score movement alongside the number of changing decisions. 

Reversing A/B option order exposes further variability. In the archived SmolVLM-256M run, 401 of 500 semantic predictions change between typed and reversed prompts. The archived Qwen record contains 7 semantic changes among its 487 jointly observed reversed items. Each option mapping uses an independently evaluated prompt, capturing the effect of presenting the alternatives in a different configuration.

\begin{figure*}[!t]
\centering
\includegraphics[width=.91\textwidth]{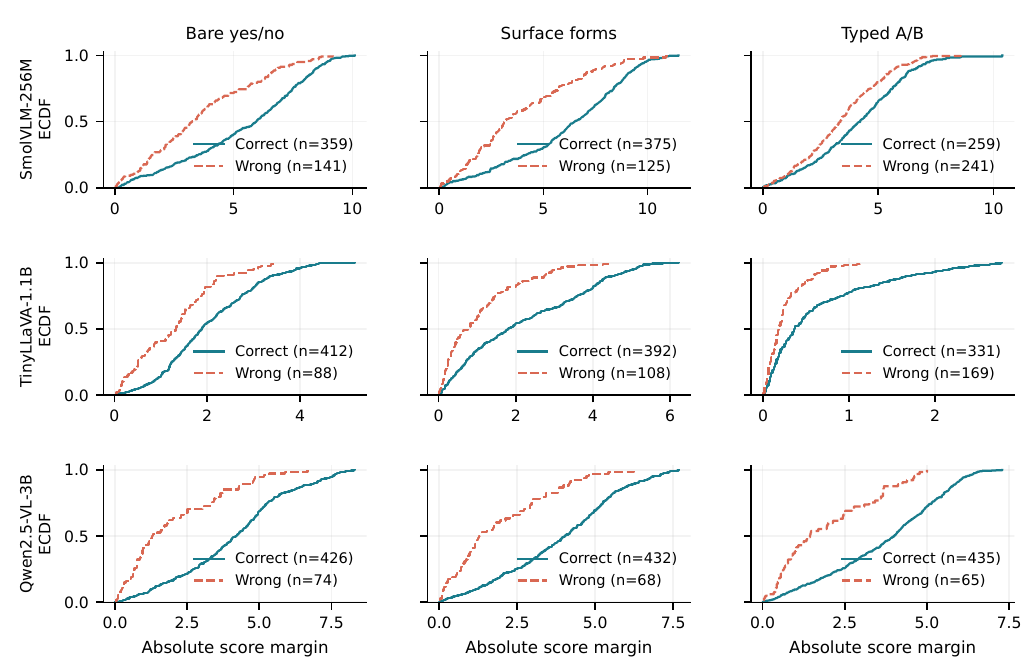}
\caption{\textbf{Correctness-conditioned absolute score-margin distributions.} Empirical cumulative distribution functions for correct and incorrect predictions across three archived POPE models and three readout interfaces. Curves show the margin distribution within each model--protocol panel; horizontal scales differ across protocols. Each panel contains 500 POPE adversarial questions.}
\label{fig:margin_distributions}
\end{figure*}

\begin{figure*}[!t]
\centering\includegraphics[width=.93\textwidth]{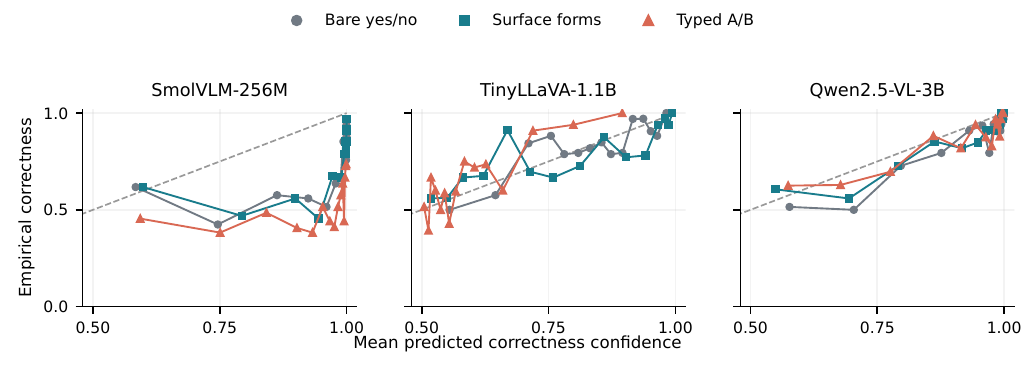}
\caption{\textbf{Reliability diagrams from archived POPE readouts.} Empirical correctness is plotted against predicted correctness confidence for bare, surface-form, and typed interfaces across three backbones. Each curve uses 500 evaluated questions and the saved 15-quantile-bin convention; the dashed line indicates ideal calibration. These historical runs are distinct from the checkpoint-pinned ECE comparison in Fig.~\ref{fig:calibration_bar}.}
\label{fig:reliability}
\end{figure*}

\subsection{Readout sensitivity extends to calibration}
Accuracy characterizes the selected label; calibration characterizes the relationship between confidence and its target event. The archived TinyLLaVA correctness ECE increases from 0.0563 under bare readout to 0.0958 under typed readout. Its positive-label ECE, using the gold-yes event, changes from 0.0469 to 0.1343. 

Correctness-margin AUROC also changes across protocols. On the archived POPE records, AUROC moves from 0.7011 to 0.6084 for SmolVLM-256M, from 0.7048 to 0.6807 for TinyLLaVA, and from 0.7927 to 0.7762 for Qwen when changing bare to typed. These differences show that answer interfaces can alter the ranking of correct and incorrect decisions by their score margins. Figure~\ref{fig:margin_distributions} presents the corresponding margin distributions for correct and incorrect predictions under each readout. Incorrect predictions generally accumulate at smaller absolute margins, and the degree of separation varies across model--protocol pairs.

\begin{figure}[!b]
\centering\includegraphics[width=\columnwidth]{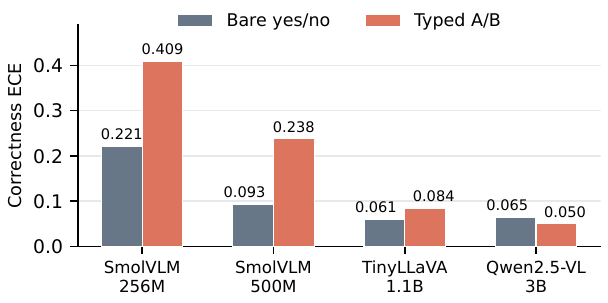}
\caption{\textbf{Calibration sensitivity in checkpoint-pinned POPE runs.} Correctness ECE (15 bins; lower is better) for bare and typed interfaces, computed on 500 matched questions per configuration. The SmolVLM-500M experiment uses its separately recorded CPU/fp32 single-tile preprocessing. Values come directly from the processed run-specific metric table.}
\label{fig:calibration_bar}
\end{figure}

The checkpoint-pinned POPE results provide an additional calibration comparison. Figure~\ref{fig:calibration_bar} compares 15-bin correctness ECE for bare and typed readout within each of the four recorded configurations. The ECE rises from 0.221 to 0.409 for SmolVLM-256M and from 0.093 to 0.238 for SmolVLM-500M. TinyLLaVA increases from 0.061 to 0.084, while Qwen decreases from 0.065 to 0.050.

Figure~\ref{fig:reliability} visualizes the corresponding diagnostic concept on the separate archived POPE runs. Each panel plots empirical correctness against predicted correctness confidence using the saved 15 quantile-bin construction. The three readout curves provide a direct visual assessment of the confidence--correctness relationship underlying ECE.

\subsection{Prompt wording and token forms}
A 500-item prompt/token matrix varies four prompt formulations and the available answer-token forms under fixed run identities. For SmolVLM-256M, the Question/Answer surface-form condition improves accuracy by $+13.2\pp$ relative to original-prompt bare scoring, with interval $[+9.4,+17.0]\pp$. The capitalized bare-token condition changes accuracy by $-10.2\pp$ with interval $[-13.2,-7.0]\pp$. 

TinyLLaVA's capitalized surface-form condition gives a $-17.2\pp$ effect with interval $[-22.4,-12.8]\pp$. Qwen's measured prompt/token cells lie between $-0.4$ and $+2.0\pp$.  The full prompt-by-token matrix is retained in the supplementary visualizations and exported result tables, preserving each individual protocol condition.

\begin{figure*}[!t]
\centering\includegraphics[width=.98\textwidth]{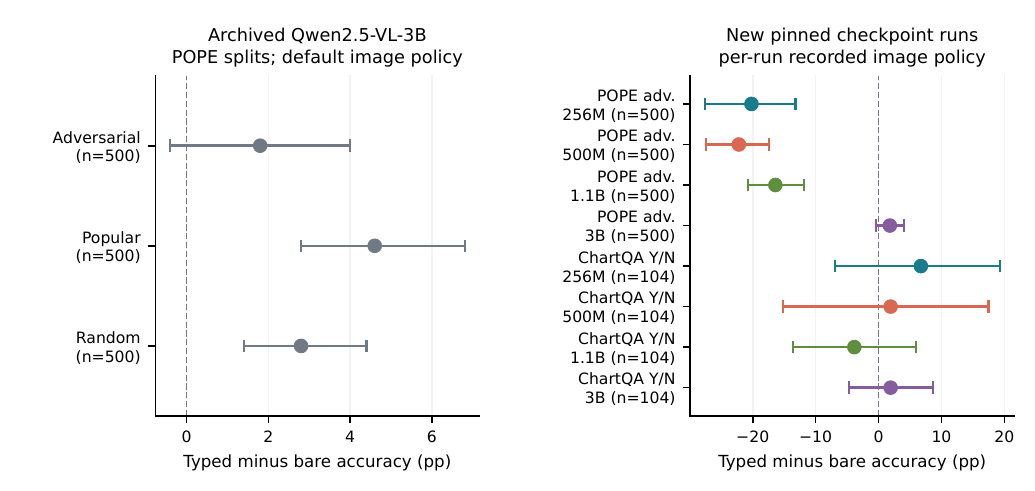}
\caption{\textbf{Readout sensitivity across visual tasks.} POPE adversarial comparisons and ChartQA native yes/no subset results are shown under their separately recorded run identities. The ChartQA study contains 104 binary questions and uses the same semantic readout definitions.}
\label{fig:cross}
\end{figure*}

\subsection{Candidate sets and option positioning}
The candidate-size study adds nested single-token foil answers to the original yes/no candidate pool at $k\in\{2,5,20,50\}$, using two membership seeds. Qwen (85.2\%), SmolVLM-500M (82.2\%), and TinyLLaVA (82.4\%) retain their baseline accuracy across the recorded candidate sizes. SmolVLM-256M remains at 64.0\% except for one $k=50$ seed at 63.8\%. For these nested inventories, an additional foil may draw the maximum score away from a gold answer, and previously selected wrong candidates remain available. This yields the construction-level monotonicity relation
\begin{equation}\label{eq:nesting}
 C_k\subseteq C_{k'}\ (k<k')\quad\Longrightarrow\quad
 \mathrm{Acc}_{k'}\le \mathrm{Acc}_{k},
\end{equation}
when foil selections are scored as invalid. 

A separate 64-item experiment changes whether the A/B options appear before or after the question. The TinyLLaVA normal-mapping suffix-to-prefix comparison yields $-23.44\pp$ with a paired interval of $[-40.63,-4.69]\pp$. Other model-specific intervals in the recorded pilot include zero.  The SmolVLM-256M CPU/fp32 versus MPS/fp16 study retains the same semantic prediction on all 64 measured questions across its four readouts, while individual score margins differ by as much as 0.136694.

\subsection{Cross-task and checkpoint analysis}
The ChartQA binary subset provides an additional setting in which the answer semantics remain yes/no while visual content changes from object-focused scenes to charts. For SmolVLM-256M, the typed-minus-bare estimate changes from $-20.2\pp$ on POPE to $+6.7\pp$ on the ChartQA subset. TinyLLaVA changes from $-16.4\pp$ to $-3.8\pp$. Qwen gives $+1.8\pp$ and $+1.9\pp$ in these two settings. All four ChartQA intervals in Table~\ref{tab:results} include zero. The cross-task contrast highlights how data composition and answer interface jointly shape the observed score.

The SmolVLM-256M and 500M checkpoints offer complementary views of the model family under their recorded device and image policies. On the binary ChartQA subset, their typed-minus-bare estimates are $+6.7\pp$ and $+1.9\pp$. On the separate POPE runs they are $-20.2\pp$ and $-22.2\pp$. Reported comparisons preserve each model's complete runtime configuration, enabling readers to reproduce the corresponding scoring interface and interpret the observed variation.

\subsection{Independent reconstruction from raw scores}\label{sec:audit}
A reliable readout comparison should be reproducible from the saved scores and candidate mappings. We independently recomputed predictions from the supplied JSONL first-token logits, run-specific token inventories, and semantic answer mappings. The replay covers all 23 recorded run identities and reconstructs 16,140 decisions. Every reconstructed decision matches its processed prediction. A second calculation from the 57,073-row per-readout table reproduces all 217 matched exported accuracy cells.

The audit also clarifies why run identity matters. Two SmolVLM-256M POPE 500-question runs produce typed-minus-bare changes of $-14.4\pp$ under CPU/fp32 and $-20.2\pp$ under MPS/fp16, with their respective preprocessing configurations. Storing each run's model revision, candidate IDs, prompt template, precision, and image policy makes such comparisons explicit. These artifacts support exact offline score replay and a transparent connection between raw model outputs and reported evaluation numbers.

\section{Conclusion}\label{sec:conclusion}
We presented \ramp{}, a readout-aware framework for systematic evaluation of multimodal large language models. Across POPE, a binary ChartQA subset, and targeted protocol controls, answer interfaces produced substantial changes in accuracy, confidence, and individual-question outcomes under fixed checkpoints and matched samples. The paired decomposition revealed large correctness transitions that can cancel in aggregate scores, while independent score replay verified the extraction of 16,140 saved predictions. \ramp{} provides a practical basis for reporting consistent, interpretable multimodal evaluation results. Future work will extend the protocol to free-form generation, larger answer spaces, and additional visual reasoning benchmarks.

\par\medskip
\section*{LLM Usage Statement}
Large language models were used for language editing. Concretely, we used GPT-6-Astra and DeepSeek-V4.1-Flash on the abstract, the introduction and other writing revision tasks to improve grammar, clarity and concision. Every AI assisted passage was reviewed and edited by the authors, who take full responsibility for the content of the paper.
\vspace{4pt}
{\small
\bibliographystyle{ieeenat_fullname}
\bibliography{references}
}
\end{document}